%% file: main_affwild2_challente.tex
\documentclass[runningheads]{llncs}
\usepackage{graphicx}
\graphicspath{ {images/} }
\usepackage{multirow}
\usepackage{tabularx}
\usepackage[british]{babel}

\usepackage{hyperref} 
\hypersetup{
    colorlinks=true
}

\usepackage{amssymb}
\usepackage{latexsym}

\usepackage{amsmath}
\usepackage{amsfonts}

\usepackage{booktabs}

\usepackage{tabularx} 

\usepackage{cleveref}
\crefname{subsection}{subsection}{subsections}

\usepackage{xstring}
\makeatletter
\AtBeginDocument{
\let\oldref\ref
\renewcommand{\ref}[1]{\IfBeginWith{#1}{fig:}%
{{\color{blue}Figure~\oldref{#1}}}%
{\IfBeginWith{#1}{tab:}{{\color{blue}Table~\oldref{#1}}}{Unsupported ref start}}}}

\title{A Multi-resolution Approach to Expression Recognition in the Wild}
\author{Fabio Valerio Massoli\inst{1*} \and Donato Cafarelli\inst{2} \and  Giuseppe Amato\inst{1} \and Fabrizio Falchi\inst{1}}

\authorrunning{F.V. Massoli et al.}
\institute{ISTI-CNR, via G. Moruzzi 1, Pisa, Italy \\
\email{\{fabio.massoli, giuseppe.amato, fabrizio.falchi\}@isti.cnr.it},\\ \and
Department of Information Engineering, Unipi, Largo L. Lazzarino 1, Pisa, Italy \\
\email{donato.caf@gmail.com}
}

\begin{document}

\maketitle

\input{1_introduction}
\input{2_related_works}
\input{3_approach}
\input{4_datasets}
\input{5_experimental_results}
\input{6_conclusions}

\section*{Acknowledgment}
We gratefully acknowledge the support of NVIDIA Corporation with the donation of the Titan V GPU used for this research.
This  work  was  partially  supported  by WAC@Lucca funded by Fondazione Cassa di Risparmio di Lucca,
AI4EU - an EC H2020 project (Contract  n.  825619), 
and upon work from COST Action 16101 ``Action MULTI-modal Imaging of FOREnsic SciEnce Evidence (MULTI-FORESEE)'', supported by COST (European Cooperation in Science and Technology).

\bibliographystyle{splncs04}
\bibliography{mybib}

\end{document}

%% file: 1_introduction.tex
\section{Introduction}
\sloppy
Facial expressions play a fundamental role in human communication. Indeed, they typically reveal the real emotional status of people beyond the spoken language. Moreover, the comprehension of human affect based on visual patterns is a key ingredient for any human-machine interaction~\cite{hci} system and, for such reasons, the task of Facial Expression Recognition (FER) draws both scientific and industrial interest. In the recent years, Deep Learning techniques reached very high performance on FER by exploiting different architectures and learning paradigms. In such a context, we propose a multi-resolution approach to solve the FER task. We ground our intuition on the observation that often faces images are acquired at different resolutions. Thus, directly considering such property while training a model can help achieve higher performance on recognizing facial expressions. To our aim, we use a ResNet-like architecture, equipped with Squeeze-and-Excitation blocks, trained on the Affect-in-the-Wild 2 dataset. Not being available a test set, we conduct tests and models selection by employing the validation set only on which we achieve more than 90\% accuracy on classifying the seven expressions that the dataset comprises. 

Since our goal is to take part to the `First Affect-in-the-Wild Challenge''~\cite{affwild2challenge}, as required by the organizers we make our code publicly available on github\footnote{\url{https://github.com/fvmassoli/affwild2-challenge.git}}.

Concerning the remaining part of the paper, we organized it as follows. In \Cref{related_works} we report several works related to the FER task, while in \Cref{approach} and \Cref{dataset} we describe our approach and the dataset we use, respectively. Moreover, we describe the experimental campaigns we perform and the corresponding model performance in \Cref{experimental_results}. Finally, in \Cref{conclusions} we conclude our work by reporting our future plans.

%% file: 2_related_works.tex
\section{Related Works} \label{related_works}


In the last two decades, several approaches have been studied to solve the FER task based on different techniques: handcraft-features, shallow and deep models. Although each technique came with its pros, generally, the DL-based ones reached the highest performance.

As an example of the handcraft-features based approches, we have Local Binary Pattern (LBP)~\cite{zhao2011facial,happy2012real}, Gabor wavelets~\cite{bartlett2003real,kotsia2008analysis}, Histogram of Oriented Gradients (HOG)~\cite{zhao2011facial,chen2014facial}, distance and angle relation between landmarks~\cite{michel2003real}, just to cite a few. Instead, in the context of shallow models-based techniques, ~\cite{suk2014real} and ~\cite{kotsia2006facial} exploited the power of the Support Vector Machines (SVMs), while~\cite{ghimire2013geometric}  proposed to use the AdaBoost classifier.

Notwithstanding the capabilities reached by the mentioned approaches, in recent years, DL-based algorithms have become state-of-the-art to tackle the FER task~\cite{Rouast_2019}.

In 2013,~\cite{icmlwin} won the ICML face expression recognition challenge by learning an SVM classifier on top of deep architectures used as backbone features extractors. Thanks to their approach, the authors achieved a 71.2\% accuracy score on the Facial Expression Recognition 2013 (FER2013)~\cite{fer2013} test set. In~\cite{Jung_2015_ICCV}, the authors used two different types of CNN to detect seven emotions on Extended Cohn-Kanade Dataset (CK+)~\cite{ck}, Oulu-CASIA~\cite{oulucasia} and MMI~\cite{mmi} datasets. 
Specifically, they combined the two models using a new integration method to boost the performance on the FER task. In~\cite{hasani2017facial}, the authors used a 3D DCNN followed by a long short-term memory (LSTM) to analyze and classify facial expressions in videos. Moreover, they also examine the proposed model's ability on cross-database classification tasks.

In~\cite{kollias2017recognition}, the authors propose a CNN-RNN architecture to valence-arousal (VA) recognition on the Aff-Wild database~\cite{zafeiriou2017aff}. Afterward, in 2019,~\cite{kollias2019deep} proposed the AffWildNet used to investigate the ability of the model to provide accurate predictions on 2D emotion labels in a variety of scenarios.

In~\cite{kollias2018multi}, the authors presented a Multi-Task learning algorithm to perform action unit (AU), expression (EX), and VA recognition, for which they exploited two different networks. Moreover, they proposed~\cite{kollias2019expression} a multi-task CNN combined with a recurrent neural network for VA and EX recognition trained on Aff-Wild2~\cite{affwild2} that they tested on ten publicly available databases.


%% file: 3_approach.tex
\section{Approach} \label{approach}
\sloppy
Usually, face images come from heterogeneous sources~\cite{massoli2020sisap}, e.g., cameras with different resolutions or different distances from the scene. Such characteristics directly impact DL models' performance on tasks such as Face Recognition (FR) by dramatically lowering their performnace~\cite{massoli2020cross}. Based on such an observation, we propose our approach grounded on the hypothesis that the images' resolution has a non-negligible impact on DL models' behavior when tested against the FER task. Specifically, we move our steps from~\cite{massoli2020cross} in which the authors explicitly take care of the multi-resolution nature of face images by designing a training technique to accommodate for such an issue adequately. 

In our work, we take inspiration from the author's training procedure, and we adapted it to our case. Specifically, we experimentally notice that we do not need any Teacher-supervised signal nor curriculum learning. Thus, we simplify the training procedure by only exploiting the double random extraction to set the final image resolution. To train the models and perform model selection, we employ the Aff-Wild2~\cite{affwild2} dataset. We refer the reader to \Cref{dataset} for a brief description of the dataset.

Our base model is a ResNet-50 architecture~\cite{he2015deep}, equipped with Squeeze-and-Excitation blocks~\cite{hu2018squeeze}, that has been pre-trained on the VGGFace2 dataset~\cite{cao2017vggface2}.  
To train our models, we use the Adam~\cite{adam} optimizer 
and set the learning rate to $1.e^{-2}$. Moreover, we set the batch size to 128, and we use data augmentation techniques to avoid overfitting. Specifically, we first resize the images to have the shortest side of 256 pixels (while keeping the original aspect ratio), then we random crop a square of 224x224 pixels, and finally, we normalize the input channels. Moreover, we apply a random grayscale conversion with a probability of 0.2. We substitute the random crop with the center one, and we remove the grayscale operation to test the model on the validation set.  

Concerning the random resolution extractions to train the models, we perform several experiments considering different ranges for the final image size concerning the multi-resolution training, with the minimum and maximum considered values being 8 and 256 pixels, respectively.

%% file: 4_datasets.tex
\section{Dataset: Affect-in-the-Wild 2} \label{dataset}
The Aff-Wild2~\cite{affwild2} dataset is the first-ever database annotated for all three main behavior tasks: VA, AU, and EX classification. Concerning the latter one, the dataset consists of 539 videos (collected from YouTube) that account for $\sim$2.6M of frames labeled considering seven expressions: neutral, anger, disgust, fear, happiness, sadness, and surprise. The annotation is made frame-by-frame by a team of seven experts. The dataset is shipped with a protocol that divides it into three non-overlapping subsets for training, validation, and test purposes. Specifically, the three partitions consist of 253, 71, and 223 videos, respectively. The cropped-aligned version of the dataset is made of images preprocessed to have a fixed resolution of 112x112 pixels.
Among the $\sim$2.6M available images, $\sim$1.2M are available for training and validation on the FER task.

As we mentioned previously, the dataset comprises seven different types of expressions with a very different cardinality. We report in \autoref{tab:affwild2_stats} the number of images for each class, both for the training and validation sets.

\begin{table}[!h]
    \centering
    \begin{tabularx}{\linewidth}{l>{\centering}X>{\centering}X>{\centering}X>{\centering}X>{\centering}X>{\centering}X>{\centering\arraybackslash}X}
    \toprule
    & \multicolumn{7}{c}{\bf Expression} \\
    & Neutral & Anger & Disgust & Fear & Happiness & Sadness & Surprise \\ \cmidrule{2-8} 
    \bf Training & 585896 & 23484 & 12497 & 11120 & 149920 & 100548 & 38564 \\ 
    \bf (\%) & (63.5) & (2.5) & (1.4) & (1.2) & (16.3) & (11.0) & (4.1) \\ \cmidrule{2-8} 
    \bf Validation & 181884 & 8003 & 5401 & 9671 & 52842 & 38534 & 22988 \\ 
    \bf (\%) & (57.0) & (2.5) & (1.7) & (3.0) & (16.5) & (12.1) & (7.2) \\ 
    \bottomrule
    \end{tabularx}
    \caption{Classes' cardinality for the Aff-Wild2~\cite{affwild2} dataset.}
    \label{tab:affwild2_stats}
\end{table}

As one can notice from \autoref{tab:affwild2_stats}, the classes are not balanced. For that reason, we leveraged a balanced cross-entropy loss to account for the class unbalance. The ``Neutral'' class represent an image where none of the other six expression has been recognized.

%% file: 5_experimental_results.tex
\section{Experimental Results} \label{experimental_results}

In this section, we report the experimental results we obtained on the Aff-Wild2~\cite{affwild2} dataset.  Since the dataset is currently employed in the Affect-in-the-Wild Challenge~\cite{affwild2challenge}, the test set's ground truth labels are not available. For such a reason, we quote the performance of our model on the validation set. Before the training, we took a small subsample of the validation set and used it for model selection purposes to avoid any bias. Subsequently, we tested the best model on the entire validation set. 
To quote our results, we use different metrics. First, we evaluate the F1-score on each class, then we summarize the overall performance of our best model across all the seven expressions by quoting the F1-score (macro-average) and the overall accuracy. Finally, we evaluate the same score as required by the Affect-in-the-Wild Challenge~\cite{affwild2challenge}, which is equal to:

\begin{equation}
     s = 0.33 \cdot \mathrm{accuracy} + 0.67\cdot \mathrm{f1\ score};
\end{equation}

where the accuracy and the f1 score are relative to the whole dataset.

We report the results in \autoref{tab:affwidl2_sc} and \autoref{tab:affwidl2_wd} concerning single class and the whole dataset, respectively 
                     
\begin{table}[!h]
    \centering
    \begin{tabularx}{\linewidth}{l>{\centering}X>{\centering}X>{\centering}X>{\centering}X>{\centering}X>{\centering}X>{\centering\arraybackslash}X}
    \toprule
    & \multicolumn{7}{c}{\bf Expression} \\ 
    & Neutral & Anger & Disgust & Fear & Happiness & Sadness & Surprise \\ \cmidrule{2-8}
    \bf F1 Score & 0.978  & 0.960 & 0.965 & 0.971 & 0.946 & 0.987 & 0.937 \\ 
    \bottomrule
    \end{tabularx}
    \caption{F1 score for each class of the Aff-Wild2~\cite{affwild2} dataset.}
    \label{tab:affwidl2_sc}
\end{table}

\begin{table}[!h]
    \centering
    \begin{tabular}{lcc} 
    \toprule
    \textbf{Accuracy} & \textbf{F1 Score} & \bf Challenge Score \\ [0.5ex] 
    & \bf (macro-average) & \\
    \midrule
     0.970 & 0.964 & 0.966 \\ 
    \bottomrule
    \end{tabular}
    \caption{Summary statistics on all the classes of the Aff-Wild2~\cite{affwild2} dataset.}
    \label{tab:affwidl2_wd}
\end{table}

From the previous tables, we can notice that our model shows promising performance on the FER task. Moreover, we acknowledge the stability of the scores among different classes even though the dataset is highly unbalanced as reported in \autoref{tab:affwild2_stats} 

%% file: 6_conclusions.tex
\section{Future Works} \label{conclusions}
In this work, we report our first experimental campaign focused FER task. We tackle such a problem by giving more representational power to our models, assuming a cross-resolution context. We observe promising results, and we are planning to submit our predictions on the test set of the Aff-Wild2~\cite{affwild2} dataset to the ``First Affect-in-the-Wild Challenge''~\cite{affwild2challenge}.